\documentclass{article}
\usepackage{ijcai22}

\usepackage{times}
\usepackage{soul}
\usepackage{url}
\usepackage[hidelinks]{hyperref}
\usepackage[utf8]{inputenc}
\usepackage[small]{caption}
\usepackage{graphicx}
\usepackage{amsmath}
\usepackage{amsthm}
\usepackage{booktabs}
\usepackage{algorithm}
\usepackage{algorithmic}
\usepackage{multirow}
\usepackage{enumitem}
\usepackage{color}
 
\newcommand{\eat}[1]{}
\newcommand{\eg}{{\em e.g.,~}}     
\newcommand{\ie}{{\em i.e.,~}}      

\newcommand{\tabincell}[2]{\begin{tabular}{@{}#1@{}}#2\end{tabular}}  

\title{Interpretable AMR-Based Question Decomposition for Multi-hop Question Answering}

\author{
Zhenyun Deng
\and
Yonghua Zhu\and
Yang Chen\and
Michael Witbrock\and
Patricia Riddle
\affiliations
School of Computer Science, University of Auckland, New Zealand\\
\emails
\{zden658, yzhu970\}@aucklanduni.ac.nz,
\{yang.chen, m.witbrock, p.riddle\}@auckland.ac.nz
}

\begin{document}

\maketitle

\begin{abstract}

Effective multi-hop question answering (QA) requires reasoning over multiple scattered paragraphs and providing explanations for answers. Most existing approaches cannot provide an interpretable reasoning process to illustrate how these models arrive at an answer. In this paper, we propose a Question Decomposition method based on Abstract Meaning Representation (QDAMR) for multi-hop QA, which achieves interpretable reasoning by decomposing a multi-hop question into simpler sub-questions and answering them in order. Since annotating the decomposition is expensive, we first delegate the complexity of understanding the multi-hop question to an AMR parser. We then achieve decomposition of a multi-hop question via segmentation of the corresponding AMR graph based on the required reasoning type. Finally, we generate sub-questions using an AMR-to-Text generation model and answer them with an off-the-shelf QA model. Experimental results on HotpotQA demonstrate that our approach is competitive for interpretable reasoning and that the sub-questions generated by QDAMR are well-formed, outperforming existing question-decomposition-based multi-hop QA approaches.

\end{abstract}

\section{Introduction}

Multi-hop question answering (QA) has long been a grand challenge in Artificial Intelligence \cite{chen2020hybridqa,Yang0ZBCSM18}. General solutions could benefit from interpretability, i.e., providing evidence for the answer by mimicking human reasoning in answering a multi-hop question. Such evidence is usually embodied in key information scattered in multiple paragraphs of the text. To endow a multi-hop QA model with better interpretability, it is desired to capture clues from the given questions and subsequently extract correct pieces of evidence based on these clues.

Currently, most multi-hop QA approaches achieve post-hoc-interpretable reasoning by constructing an auxiliary model to provide explanations for an existing QA model, \eg \cite{qiu2019dynamically,tu2020select,fang2019hierarchical} perform interpretable multi-hop reasoning using a supporting-evidence prediction model. As a result, methods of this type usually cannot employ a context-specific reasoning process for answering multi-hop questions; this disconnect between reasoning and explanation lessens  the extent to which humans can understand the behaviour of the machine reasoning models and therefore how these models might be improved. In contrast, methods based on intrinsically interpretable reasoning may better help humans to understand and trust the mechanism of a QA system using multi-hop reasoning, because they construct a self-explanation system and incorporate interpretability directly into that system.

\begin{figure}[!t]
	\begin{center}
		{\scalebox{0.60} {\includegraphics{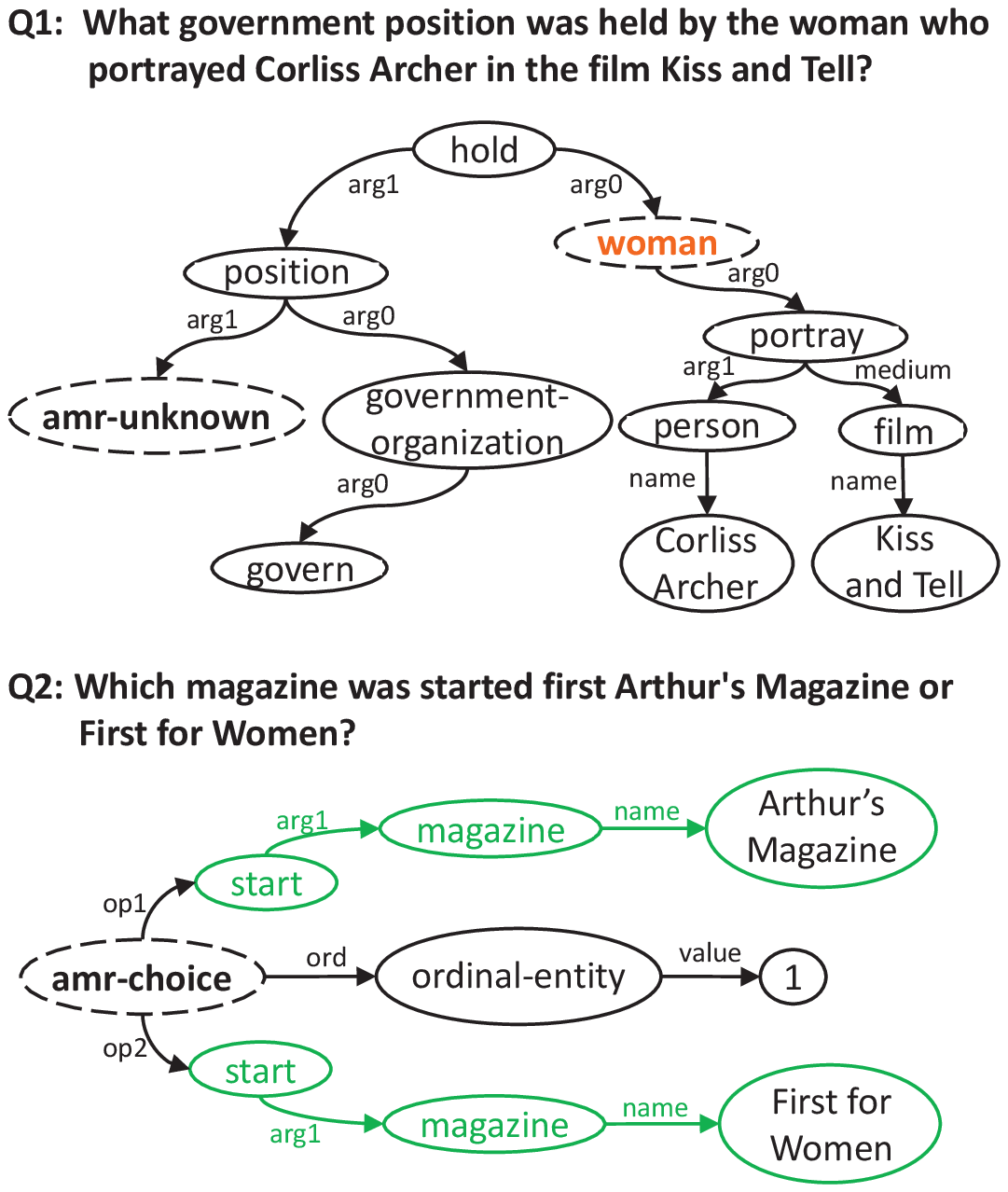}}}
		\vspace{-3mm}
		\caption{\footnotesize Examples of delegating the complexity of understanding multi-hop questions to AMR parsers. Q1 is a bridging multi-hop question; these can be interpreted by identifying the secondary ``unknown" variable \textit{woman}. Q2 is a comparison question; these can be interpreted by capturing the longest identical path.}
		\label{fig1}
	\end{center}
	\vspace{-5mm}
\end{figure}

Divide-and-conquer is a promising strategy for intrinsic interpretability, in which complex multi-hop questions are decomposed into simpler sub-questions that can be answered by an off-the-shelf QA model. The simplified sub-questions may help humans to understand the logic of complex multi-hop questions better, and also to diagnose where QA models fail. Since annotations for decomposition are expensive and time-consuming, existing studies have trained a question decomposition (QD) model using a combination of hand-crafted labelled examples and a rule-based algorithm, or have trained a pseudo decomposition model to extract spans of text as sub-questions \cite{min2019multi,perez2020unsupervised}. However, the resulting sub-questions are often ungrammatical, may not even be explicit questions, or require additional language processing, leading to errors in downstream prediction tasks. 

In a multi-hop question, predicate-argument structure plays a key role in understanding its meaning. 
Utilizing predicate-argument structures to help understand natural language questions has become a common approach \cite{rakshit2021asq}. In the explicit graph structure of AMR, the predicate-argument structure appears in the form of nodes. Recently, AMR parsers have made great progress in reducing the complexity of understanding natural language questions by improving their semantic graph representations \cite{ribeiro2020investigating}, benefiting downstream tasks such as summarization \cite{liao2018abstract} and question answering \cite{xu2021dynamic}. Furthermore, AMR-to-Text generation models have been developed in recent years that produce well-formed text by encoding the AMR graph as an input to pretrained language models \cite{bevilacqua2021one,ribeiro2021structural}.

Motivated by this, we propose to perform question decomposition based on AMR for multi-hop QA. Our method is designed for four goals: i) delegating the complexity of understanding multi-hop questions to AMR parsing; as shown in Figure \ref{fig1} we understand multi-hop questions better by identifying the intermediate unknown variable or longest identical path; ii) segmenting the AMR graph according to the required type of reasoning; iii) generating sub-questions by AMR-to-Text generation and sorting them into a logical order, \eg based on primary and secondary unknowns; iv) answering sub-questions and providing an interpretable reasoning process that can form a basis for explanation.

The contributions of this paper are summarised as follows:
\begin{itemize} \setlength\itemsep{2pt}
    \item We transform the task of decomposition of the multi-hop question into the task of segmentation of the corresponding AMR graph.
    
    \item We propose two different AMR graph segmentation methods for multi-hop QD according to the reasoning type required: 1) unknowns-based and 2) path-based graph segmentation.
    
    \item We propose a pipeline-based modular method that integrates multiple modules, \ie AMR parsing, AMR graph segmentation and AMR-To-Text generation, for multi-hop QD and multi-hop QA, which may help humans to better understand the reasoning behaviors of an interpretable multi-hop QA model. 
\end{itemize}

\begin{figure*}
	\begin{center}
		{\scalebox{0.52} {\includegraphics{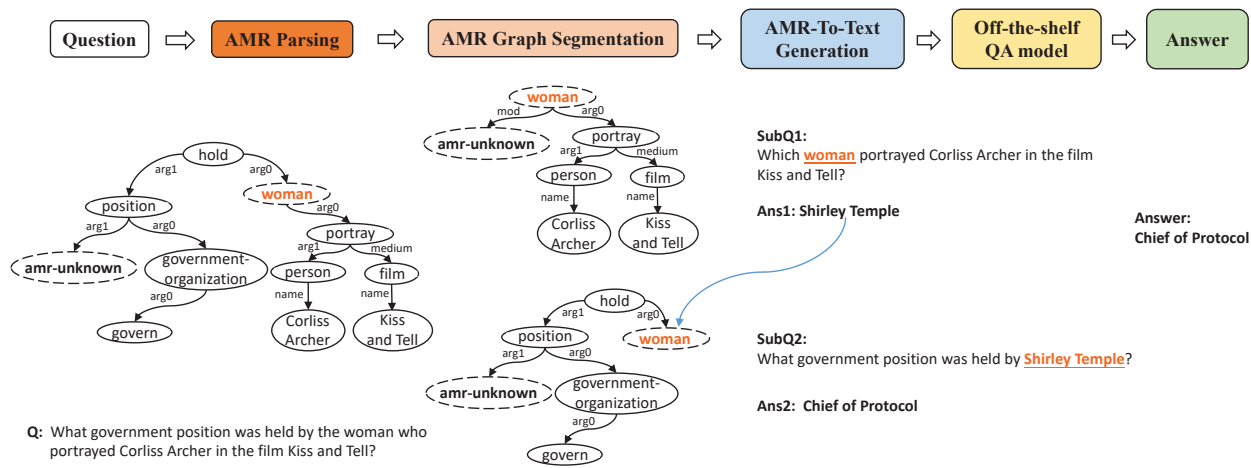}}}
		\caption{\footnotesize
		An overview of our proposed QDAMR framework for multi-hop QA. QDAMR first parses question Q, \textit{What government position was held by the woman who portrayed Corliss Archer in the film Kiss and Tell?}, into an AMR graph; it then segments the graph based on the secondary unknown variable \textit{woman}; next it applies AMR-to-Text generation to generate well-formed sub-questions, subQ1 and subQ2; after that it uses off-the-shelf QA to predict the answer of the secondary sub-question subQ1; and finally QDAMR substitutes that answer into the unknown \textit{woman} in the primary sub-question to predict the final answer, again via off-the-shelf QA.}
		\vspace{-3mm}
		\label{framework}
	\end{center}
	\vspace{-3mm}
\end{figure*}

\section{Related Work}
\noindent \textbf{Intrinsic Interpretability.}
Intrinsically interpretable QA models incorporate interpretability directly into the structure of QA systems, making them self-explanatory. Recent studies have used divide-and-conquer, simplifying multi-hop questions to achieve self-explainable reasoning. DecompRC \cite{min2019multi} decomposes complex questions into several sub-questions and uses a scoring model to jointly select the final answer and the most appropriate decomposition. OUNS \cite{pan2020unsupervised} creates a pseudo-decomposition model with unsupervised seq2seq learning to map a hard question to many simple questions, and builds a recomposition model that combines answers of those simple questions to obtain the final answer. Unlike the above methods, QDAMR decomposes a multi-hop question based on the high-level semantic relations between the concepts in its AMR parse graph, and answers sub-questions in a logical, and potentially explanatory, order.

\smallskip
\noindent \textbf{Post-hoc Interpretability.}
Post-hoc-interpretable QA models provide explanations for multi-hop reasoning by creating second (supporting evidence) models, which fall into two categories. In one, the supporting evidence prediction model is constructed independently: AISO \cite{zhu2021adaptive} defines three types of retrieval operations to find the missing evidence at each reasoning step. In the other category a multi-task model achieves answer prediction and supporting evidence prediction simultaneously. HGN \cite{fang2019hierarchical}, Entity-GCN \cite{de2018question} and DFGN \cite{qiu2019dynamically} fuse multi-granularity information in a graph, and then apply a multi-task prediction model over the graph to achieve both answer and evidence prediction.

\smallskip
\noindent \textbf{AMR for Multi-hop QA.}
AMR parsers have made great progress  in recent years as a source of explicit graph structure for conducting symbolic reasoning \cite{mitra2016addressing}. AMR-SG \cite{xu2021dynamic} constructs an AMR semantic graph from valid evidence items, and reasons over it to obtain interpretability. NSQA \cite{kapanipathi2021leveraging} proposes a Neuro-Symbolic QA system based on AMR that forms logical queries and applies a neuro-symbolic reasoner to predict the final answer. Unlike the above forms of symbolic reasoning, we use AMR to convert the multi-hop question to symbolic form in support of question decomposition and identification of intermediate unknowns.

\section{Method}

\subsection{Overview}
In this section, we describe interpretable question decomposition based on AMR, for multi-hop QA. As illustrated in Figure 2, the pipeline of our proposed QA system consists of four modules: i) AMR Parsing: utilizes AMR parsers to transform the multi-hop questions into an AMR graph; ii) AMR graph segmentation: segments the AMR graph into simpler AMR sub-graphs according to the reasoning type of the question;  iii) Sub-question generation: applies AMR-to-Text generation model to sub-graphs to generate sub-questions; iv) Single-hop QA model: answers sub-questions with off-the-shelf QA models.

\subsection{AMR Parsing}
Multi-hop QA requires multiple steps of reasoning over multiple scattered paragraphs to arrive at an answer. In order to help humans understand how these models make decisions, it is useful to capture intermediate unknown variables in the multi-hop question. To do this, we delegate the complexity of understanding the multi-hop question to an AMR graph. As shown in Figure \ref{framework}, we parse a multi-hop question Q into an AMR graph, where nodes represent concepts and edges represent relations in between. An \textit{amr-unknown} node denotes a primary unknown concept that represents the answer to the given question Q. In the example of Figure \ref{framework}, \textit{amr-unknown} is an AMR \textit{position}  which represents the type of the answer. Furthermore, the AMR graph annotations on Q can also help identify intermediate unknown variables needed for multi-hop reasoning, \eg \textit{woman} is a secondary unknown. 

We exploit transfer learning, using a pretrained model based on BART \cite{lewis2019bart} for AMR parsing to a linearized AMR graph \cite{bevilacqua2021one}. Its advantage is that it overcomes the data sparsity issue in seq2seq-based methods. Since the linearized graph output is fully graph-isomorphic, we encode the question into a graph without losing adjacency information.  

\begin{algorithm}[tb]
    \small
    \caption{: Question Decomposition Based on AMR}
    \label{alg:algorithm}
    \textbf{Input}:  Multi-hop Question Q \\
    \textbf{Output}: Bridging/Intersection question: q$_1$, q$_2$; \\ Comparison question: q$_1$, q$_2$, q$_3$
    \begin{algorithmic}[1] 
    
    \STATE Linearized AMR Graph  G $:=$\textbf{AMR Parsing}(Q)
    \STATE Q$_{type}$ := \textbf{getReasoningType}(G)
    \IF {Q$_{type}$ is bridging}
    \STATE secUnkNode $:=$ \textbf{getSecondaryUnknows}(Q, G)
    \STATE G$_1$, G$_2$ $:=$ \textbf{unkBasedSegmentation}(G, secUnkNode)
    \ELSE 
    \STATE longestPath $:=$ \textbf{getLongestIdenticalPath}(G)
    \STATE G$_1$, G$_2$ $:=$ \textbf{pathBasedSegmentation}(G, longestPath)
    \ENDIF
    \STATE q$_1$ $:=$ \textbf{AMR-to-Text Generation}(G$_1$)
    \STATE q$_2$ $:=$ \textbf{AMR-to-Text Generation}(G$_2$)
    
    \IF {Q$_{type}$ is bridging}
    \STATE q$_1^{\rm Ans}$ $:=$ \textbf{off-the-shelf QA}(q$_1$)
    \STATE G$_2^{'}$ $:=$ \textbf{replace}(G2, q$_1^{\rm Ans}$, secUnkNode)
    \STATE q$_2$ $:=$ \textbf{AMR-to-Text Generation}(G$_2^{'}$)
    \ELSIF{Q$_{type}$ is intesection}
    \STATE Continue
    \ELSE 
    \STATE op $:=$ \textbf{findOperation}(Q)
    \STATE q$_1^{\rm Ans}$, q$_2^{\rm  Ans}$ $:=$ \textbf{off-the-shelf QA}(q$_1$, q$_2$)
    \STATE q$_3$ $:=$ \textbf{constructOpQuestion}(op, q$_1^{\rm  Ans}$, q$_2^{\rm  Ans}$)
    \ENDIF
    \IF {Q$_{type}$ is bridging or intersection}
    \STATE \textbf{return} q$_1$, q$_2$
    \ELSE
    \STATE \textbf{return} q$_1$, q$_2$, q$_3$
    \ENDIF
    \end{algorithmic}
\end{algorithm}

\subsection{AMR Graph Segmentation}
The key challenge of question decomposition is that it is difficult to obtain annotations for the decomposition. We therefore propose to transform decomposition of the multi-hop question into a corresponding AMR graph segmentation operation. The advantage of AMR graph segmentation is that it can be done without linguistic expertise and without training. Multi-hop questions can be divided into the following categories according to the reasoning type: bridging, intersection and comparison. Thus, we propose two graph segmentation methods across the three reasoning types, \ie unknown-based and path-based AMR graph segmentation, which we describe in the following subsection.
\smallskip

\noindent \textbf{Unknowns-based Graph segmentation.} 
Bridging questions require asking for the answer to an entity that is not explicitly mentioned. Therefore, the key to answering the question is how to identify the secondary unknown entity. According to  \cite{rakshit2021asq}, each predicate-argument relationship at the sentence level can be expressed as a question-answer pair. Motivated by this, we propose using unknowns-based graph segmentation for bridging questions. We first extract all predicate nodes, then take subject nodes connected to the predicate nodes as candidate unknowns, and finally segment the AMR graph based on subject nodes. This is done by line 1-5 in Algorithm 1. In the example of Figure \ref{framework}, \textit{portray} is the predicate node and \textit{woman} is the corresponding subject node which is the secondary unknown in the question Q. 

\vspace*{0.5\baselineskip}
\noindent \textbf{Path-based Graph segmentation.} 
Intersection questions and comparison questions share the commonality that they both have two parallel conditions or entities. Intersection questions require asking for an entity that satisfies multiple conditions (\textit{``Are both Coldplay and Pierre Bouvier from the same country?"}), and comparison questions require comparing a property of two entities (\textit{``Who is older, Annie Morton or Terry Richardson?"}). Thus, we propose a path-based method to capture these parallel conditions or entities by retrieving the longest identical paths in the AMR. Specifically, we take the linearized AMR graph as a sequence of symbols, made up of concepts and relations, and then view the longest identical sequences as identifying parallel conditions or entities (line 7 in Algorithm 1). We demonstrate this longest path retrieval on question Q2 in Figure \ref{fig1}. The following is the longest identical path; fields prefixed with a colon represent the relationship between concepts, \eg \rm{:arg} and \rm{:name}.

\begin{center}
$\rm {{start{\rightarrow}:arg{\rightarrow}magazine{\rightarrow}:name{\rightarrow}<Entity>}}$
\end{center}

\begin{figure}
	\begin{center}
		{\scalebox{0.6}
		{\includegraphics{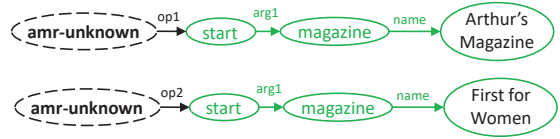}}}
		\vspace{-4mm}
		\caption{\footnotesize An example of path-based AMR graph segmentation by disconnecting instances of the longest identical path in turn.}
		\label{fig3}
	\end{center}
	\vspace{-5mm}
\end{figure}

Finally, we generate sub-graphs by disconnecting the path instances in turn, as shown in Figure \ref{fig3}.

\subsection{Sub-question Generation}
After graph segmentation, we apply an AMR-to-Text generator with the same architecture as the AMR parser to generate sub-questions from AMR sub-graphs \cite{bevilacqua2021one}. This model first transforms the graph into a sequence of symbols using a linearization technique, and then feeds it into the pretrained language model BART to obtain the sub-question. Similar to AMR parsing, the full graph-isomorphic linearization technique encodes the graph into a sequence of symbols without losing adjacency information. This helps make the generated text fluent. Especially, for comparison questions, we apply a single discrete operation \cite{min2019multi} over the answers of sub-questions to generate the third sub-question (line 21 in Algorithm).

\subsection{Single-hop Question Answering}
To achieve interpretable reasoning, we first determine the order of sub-questions before answering them, because the answer to one sub-question may be part of another, \eg the answer to question Q1 \textit{Shirley Temple} is equivalent to \textit{woman} in AMR graph of Figure \ref{framework}. Thus, we use different strategies to answer sub-questions for different reasoning types. For bridging questions, we first predict answers to sub-questions with secondary unknowns, then substitute the answer for the secondary unknowns in the primary sub-question, and finally take the answer of this primary sub-question as the final answer. For intersection questions, we do not need to identify the order of sub-questions because sub-questions are parallel, and the final answer is the intersection of answers of sub-questions. Similarly, the final answer of the comparison question is the result of applying some discrete operation over the answers of sub-questions \cite{min2019multi}. 

Given ordered sub-questions, we train a single-hop QA on SQUAD \cite{rajpurkar2016squad} and on the new single-hop QA dataset. The dataset consists of single-hop QA pairs constructed by \cite{pan2020unsupervised} on HotpotQA. Once trained, we use the resulting single-hop QA system as the off-the-shelf QA model and no longer fine-tune the model on multi-hop questions. In addition, we compute a forward pass on each sentence with sub-questions, if the sentence contains the answer of sub-questions, we take it as supporting evidence. In the end, sub-question-evidence pairs are used to support the interpretability of the single-hop QA model.

\eat{; otherwise, we predict ``no evidence".}

\section{Experiments}
We empirically evaluate our QDAMR against state-of-the-art QD-based QA models on the HotpotQA \cite{yang-etal-2018-hotpotqa}.

\subsection{Data and Setup}
\noindent \textbf{Dataset.} We evaluate our model under the distractor setting of HotpotQA, which consists of questions and a collection of 10 paragraphs. Each question is originally annotated with either of two reasoning types: bridging and comparison. For the sake of fine-grained type annotations, all our methods further separate out a third type of question, intersection, from the bridging type, based on specific AMR concepts.

In addition, following \cite{pan2020unsupervised}, we construct new single-hop QA pairs from HotpotQA using the following procedures: i) for bridging questions, we first find an entity \textit{a} that links the gold-standard evidence, then use the pretrained transformer model T5 \cite{raffel2019exploring} to generate a question \textit{Q} with \textit{a} as the answer from the gold-standard evidence, and finally take \textit{(Q, a)} as a new QA pair; ii) for intersection and comparison questions, we first find potential comparison entities \textit{e$_1$/e$_2$} from the gold-standard evidence, then generate a question \textit{Q$_1$/Q$_2$} that contains a specific entity \textit{e$_1$/e$_2$}, and finally take \textit{(Q$_1$, e$_1$)} and \textit{(Q$_2$, e$_2$)} as a new QA pair.

\vspace*{0.5\baselineskip}
\noindent \textbf{Baselines.} 
We compare our method with other QD-based QA on HotpotQA. For a fair comparison, we use the BERT-based QA from \cite{min2019multi} as the off-the-shelf QA.

\begin{itemize}[leftmargin=*]
    \item {\bf DecompRC} \cite{min2019multi} trains a pointer model to identify split points in a question; these are subsequently used to compose sub-questions for each reasoning type. 
    \item {\bf OUNS} \cite{perez2020unsupervised} creates a noisy pseudo-decomposition for each multi-hop question, and then trains a decomposition model with unsupervised seq2seq learning to improve the pseudo-decomposition. 
    \item  {\bf QDAMR}: QDAMR$^{^{\rm T5}}$ denotes the T5-based transformer model used for AMR parsing and AMR-to-Text generation; QDAMR$^{^{\rm SP}}$ denotes that we use Spring-based transformer model for AMR parsing and AMR-to-Text generation;
    QDAMR${_{_{\rm RoBERTa}}^{^{\rm SP}}}$ denotes a finetuned RoBERTa-based QA model that is used as the off-the-shelf QA model.
\end{itemize}
\smallskip

\noindent \textbf{Setup.}
\eat{Each of involved methods separates the questions in the type of bridging in a different manner. For both fairness and comprehension in comparisons, we conduct two sets of experiments by adopting the separating results of the DecompRC (as a representative of baselines) and our QDAMR, respectively.}
Each of the tested methods separates the questions by type in a different manner. For both fairness and comprehensiveness in comparisons, we conduct two sets of experiments, each using either the DecompRC and QDAMR strategies with each tested system. 
Specifically, DecompRC uses the different question decomposition methods per reasoning type to answer the multi-hop question, and then uses a decomposition scorer to select the best answer as the final result. The reasoning type that yields the best result is returned as the reasoning type of the multi-hop question.  While, our QDAMR first identifies the reasoning type of the question (see line 2 in Algorithm 1), and then uses the corresponding AMR-based decomposition method to answer the question.
\smallskip

\noindent\textbf{Performance Metrics.} Exact Match (EM) and partial match (F1) between the prediction and the ground truth answer are used as performance metrics \cite{yang-etal-2018-hotpotqa}.

\begin{table}
\small
	\centering
		\eat{\caption{\footnotesize The analysis of multi-hop QA model based on QD.}}
		\begin{tabular} {c c c c c} \toprule
			\multirow{2}{*}{\tabincell{c}{Decomp\\ Method}}  &\multicolumn{2}{c}{Annota$_{_{\small \rm QDAMR}}$} &\multicolumn{2}{c}{Annota$_{_{\small \rm DRC}}$}\\ \cmidrule(lr){2-3}   \cmidrule(lr){4-5} 
			 &EM &F1 &EM &F1  \\\midrule
			 DecomRC  & 54.82 & 68.23 & 55.12 &69.99 \\
			 OUNS     & 58.74 & 73.40 & 58.74 &73.40 \\ 
			 QDAMR$^{^{\small \rm T5}}$ & 62.24 & 77.19 &	60.07 &75.34 \\ \midrule
			 QDAMR$^{^{\small \rm SP}}$ & 64.57 & 78.44 & 63.13 & 77.61 \\
			 QDAMR${_{_{\small \rm RoBERTa}}^{^{\small \rm SP}}}$ &\bf67.13	& \bf81.17 & \bf66.83 & \bf80.12\\ \bottomrule
		\end{tabular}
		\caption{\footnotesize Results for QD-based multi-hop QA models on the dev set of HotpotQA. {Annota$_{_{\small \rm QDAMR}}$} denotes that the reasoning types of multi-hop questions are annotated by QDAMR. {Annota$_{_{\small \rm DRC}}$} denotes that the reasoning types of questions are annotated by the baseline DecompRC. RoBERTa denotes that we use a finetuned RoBERTa as the QA model. QDAMR achieves state-of-the-art results under both annotations. }
	\label{tab1}
\end{table}

\begin{table}
    \small
	\centering
	\begin{tabular} {c c c c c} \toprule
		\tabincell{c}{Decomp\\ Method}  &\tabincell{c}{GPT2 $\downarrow$ \\ NLL } &\tabincell{c}{\%Well- $\uparrow$ \\formed} &\tabincell{c}{Edit $\downarrow$ \\ Dist} &\tabincell{c}{Length $\downarrow$ \\Ratio}  \\  \midrule 
		DecompRC &6.04 &32.6 &7.08 &1.22 \\
		OUNS     &5.56 &60.9 &5.96 &1.08 \\
		QDAMR    &\bf5.37 &\bf68.2 &\bf5.19 &\bf1.06 \\ \bottomrule
	\end{tabular}
	\caption{\footnotesize Quality Analysis of the sub-questions. GPT2 NLL reflects the fluency of the question.\%Well-formed,  Edit distance and Length Ratio reflect the wellformedness of the sub-question. $\downarrow$ means lower is better, and $\uparrow$ means higher is better.}
	\label{tab2}
\end{table}

\begin{table}
    \small
	\centering
	\begin{tabular} {c| c| c| c} \toprule
		\tabincell{c}{AMR\\ Parsing}  &\tabincell{c}{AMR-to-Text\\ Generation} &EM &F1  \\  \midrule 
		CAIL &T5 &58.89  &74.34\\
		T5 &T5 &62.24 & 77.19 \\
		Graphere &T5 &63.36 &77.83 \\ \midrule
		Spring  &Spring &\bf67.13	&\bf81.17 \\ \bottomrule
	\end{tabular}
	\caption{\footnotesize Analysis of the combination of AMR parsing and AMR-to-Text generation. The combination of Spring-based AMR parsing and Spring-based AMR-to-Text achieves the best EM/F1 score. }
	\eat{The remaining four cases are not included since in these cases the output of AMR Parsing is incompatible with the input of the AMR-to-Text generator.}
	\vspace{-2mm}
	\label{tab3}
\end{table}

\subsection{Main Results}
Table \ref{tab1} shows how the question decomposition method affects performance using the two question-type identification methods. Under the same setting, using a BERT-based QA model, our QDAMR models yield substantial improvements, where QDAMR$^{^{\rm \tiny SP}}$ respectively improves EM/F1 by 9.75/10.21 and 5.83/5.04 in {Annota$_{_{\tiny \rm QDAMR}}$}, and by 8.01/7.62 and 4.39/4.21 in {Annota$_{_{\tiny \rm DRC}}$}, compared to DecompRC and OUNS. This demonstrates the effectiveness of our model's use of AMR to decompose multi-hop questions. Moreover, the performance under the {Annota$_{_{\tiny \rm QDAMR}}$} setting consistently outperforms that under the {Annota$_{_{\tiny \rm DRC}}$} setting, validating improved effectiveness identifying reasoning type over the DecomRC baseline. By fine-tuning RoBERTa on generated single-hop QA pairs and SQUAD, our model shows further improvements on EM/F1 with 2.56/2.73 in {Annota$_{_{\tiny \rm QDAMR}}$} and 3.7/2.59 in  {Annota$_{_{\tiny \rm DRC}}$}, compared to the QDAMR$^{^{\tiny \rm SP}}$, and a substantial improvement over both the DecompRC and OUNS baselines. In the following ablation studies section, we analyse the quality of the sub-questions and sources of performance gain.

\begin{table*}
	\small
	\centering
	\begin{tabular} {|c|c|c|c|} \hline
	  	\multicolumn{4}{|c|} {\textbf{Comparison Question}: Who is older, Annie Morton or Terry Richardson? \quad \textbf{Answer: Terry Richardson}} \\ \hline
	  	Methods &DecompRC &OUNS &QDAMR \\ \hline
	  	subQ1 &Who is older?  & Who is Annie Morton?	 &How old is Terry Richardson? \\
	  	Ans1 &(Annie Morton) &(American model) &(August 14, 1965)  \\
	  	
	  	subQ2 &Who is Annie Morton or Terry Richardson?  &When was Terry Richardson born?  & What was Annie Moore's age? \\
	  	Ans2 &(Annie Morton) &(26 July 1999) & (October 8, 1970) \\
	  	subQ3 &Which is smaller (Ans1)(Ans2)? &$-$ &Which is smaller (Ans1)(Ans2)? \\ 
	  	Final Ans  &Annie Morton &26 July 1999 &\textbf {Terry Richardson} \\ \hline
	  	
	  	\multicolumn{4}{c}{} \\ \hline
	  
		\multicolumn{4}{|c|}{\textbf{Intersection Question}: Are both Coldplay and Pierre Bouvier from the same country?	\quad			 \textbf{Answer: No}} \\  \hline
		Methods &DecompRC &OUNS &QDAMR \\ \hline
		subQ1 & Are both coldplay?   &Where are Coldplay and Coldplay from?  &From what country is ColdPlay?   \\ 
		Ans1 &(British rock band) &(British) &(British) \\
		
		subQ2 & Are pierre bouvier from the same country?  &What country is Pierre Bouvier from?  &Where is Pierre Bouvier from? \\
		Ans2 &(Canadian) &(Canadian) &(Canadian)  \\
	    operation &{$Intersection$}(Ans1,Ans2) &$-$ &{$Intersection$}(Ans1,Ans2) \\ 
	  	Final Ans  &\textbf {No} &British &\textbf {No} \\ \hline
		
	\end{tabular}
	\caption{\footnotesize Examples of sub-questions generated by the evaluated decomposition methods. Our method is competitive in both the accuracy of the final answer and the quality of sub-questions.}
	\vspace{-2mm}
	\label{casestudy}
\end{table*}

\subsection{Quality of Sub-questions}
We evaluate the quality of sub-questions from two perspectives: fluency and wellformedness. We measure fluency using the GPT-2 \cite{radford2019language} negative log-likelihood (NLL). For wellformedness, we use three metrics from OUNS: i) the proportion of well-formed sub-questions \cite{faruqui2018identifying}; ii) the token Levenstein (edit) distance between the multi-hop question and its generated sub-questions; and iii) the ratio between the length of the multi-hop question and the length of the sub-question.

As shown in Table \ref{tab2}, QDAMR outperforms  DecompRC and OUNS on these four metrics; QDAMR achieves the best result on GPT-2 NLL score with 5.37 (lower is better). For the proportion of well-formed questions, QDAMR achieves the best score with 68.2, indicating that sub-questions generated by our model are more grammatical. The lowest Levenstein edit distance score indicates that the sentence structure of our generated sub-questions is more similar to that of the original multi-hop question.

\subsection{Ablation Studies}
Next, we verify the effectiveness of two modules, AMR paring and AMR-to-Text generation, in our proposed framework and analyse the performance of different decomposition methods across three reasoning types.

\paragraph{Effectiveness of AMR Parsing and AMR-to-Text Generation.} Since we delegate the complexity of understanding multi-hop questions to the process of turning it into an AMR graph, and the generation of a well-formed sub-question to the process of AMR-based text synthesis, it is necessary to select a good combination of AMR parsing and AMR-to-Text generation. To this end, we use four AMR parsers (CAIL \cite{cai2020amr}, Graphere \cite{hoang2021ensembling}, T5 and SPRING \cite{bevilacqua2021one}) to generate the AMR graph from a multi-hop question, and we generate sub-questions from AMR sub-graphs using two AMR-to-Text generation methods (T5 and SPRING) because CAIL and Graphere do not support AMR-to-Text generation. In Table \ref{tab3}, we observe that sub-questions generated by the integration of the Spring-based AMR parsing and Spring-based AMR-to-Text generation subsystems achieve the best EM/F1 scores. This indicates the value of the Spring architecture in which full graph isomorphism allows for encoding and decoding without losing adjacency information, resulting in high fluency of generated sub-questions.
\begin{table}
    \small
	\centering
	\begin{tabular} {c c c c c c c} \toprule
		\multirow{2}{*}{\tabincell{c}{Decomp\\ Method}}  &\multicolumn{2}{c}{bridge} &\multicolumn{2}{c}{intersec} &\multicolumn{2}{c}{comparison} \\ \cmidrule(lr){2-3}   \cmidrule(lr){4-5} \cmidrule(lr){6-7}
		&EM &F1 &EM &F1 &EM &F1 \\\midrule
		DecomRC & 55.24 & 71.53 & 54.55 & 69.29 & 52.81 & 63.44\\
		OUNS    & 66.41 & 80.84 & 66.93 & 81.07 & 65.62 & 79.43\\ 
		QDAMR   & \bf69.45 & \bf82.35 & \bf66.98 & \bf81.15 & \bf66.02 & \bf80.24\\ \bottomrule
	\end{tabular}
	\caption{\footnotesize Performance analysis of decomposition across reasoning types. QDAMR-based QA model substantially outperforms the other QD-based QA on bridging questions, and may have a small performance advantage over OUNS for the other reasoning types.}
	\vspace{-3mm}
	\label{tab4}
\end{table}
\smallskip

\noindent \textbf{Decomposition Performance across Reasoning Types.}
We compare the performance of the three decomposition methods across reasoning types. As shown in Table \ref{tab4}, QDAMR outperforms other methods on all three tested reasoning types. 
We observe that the EM/F1 scores of our QDAMR on bridging questions (69.45/82.35) are improved by 3.04/1.51 and 14.21/10.82 compared with OUNS and DecompRC, respectively, indicating that unknowns-based decomposition can successfully identify the primary and secondary unknowns of the multi-hop question to generate well-formed sub-questions. For intersection and comparison questions, QDAMR still has a significant margin over DecompRC and slightly outperforms OUNS. This suggests that explicit path-based decomposition may hold advantages over the implicit pointer-based decomposition of DecompRC. The examples in Table \ref{casestudy} illustrate how decomposition in QDAMR works across reasoning types that require additional combination operations (intersection and comparison). We also observe that sub-questions generated by QDAMR are more consistently well-formed. In addition, our proposed pipeline-based modular method is a self-explanatory system, 
which directly incorporates interpretability into the structure of its QA pipeline; the following are the interpretable reasoning processes applied for each of the reasoning types, \textbf{B}ridging, \textbf{I}ntersection and \textbf{C}omparison:
\begin{align*}
\small
  & {\textbf{B}: \rm {Q{\rightarrow}SubQ1{\rightarrow}Ans1{\rightarrow}SubQ2{\rightarrow}Ans}}\\
  & {\textbf{I}: \rm  {Q{\rightarrow}(SubQ1, SubQ2){\rightarrow}intersec(Ans1, Ans2){\rightarrow}Ans}}\\
  & {\textbf{C}: \rm  {Q{\rightarrow}(SubQ1, SubQ2){\rightarrow}(Ans1,Ans2){\rightarrow}SubQ3{\rightarrow}Ans}}. 
\end{align*}

\section{Conclusion and Future Work}
We have demonstrated use of question decomposition based on the AMR semantic representation for multi-hop QA, using an intrinsically interpretable framework to incorporate interpretability directly into the system structure. The complex task of multi-hop question interpretation is delegated to AMR parsers. These parsers produce AMR Graphs to which two segmentation methods are applied, \ie unknown-based and path-based graph segmentation, to achieve question decomposition. To generate a well-formed sub-question, we perform both AMR parsing and AMR-to-Text generation with the same architecture, which uses a fully graph-isomorphic linearization technique to complete the transformation from graph to a sequence of symbols without losing adjacency information. Experimental results demonstrate that our QDAMR system outperforms baseline question decomposition methods, both in performance of multi-hop QA and in the quality of generated sub-questions. Since our proposed graph segmentation methods are based on predicate-argument relations and parallel conditions/entities respectively, they could in principle be generalized to an unknown number of hops by identifying multiple predicate nodes or capturing multiple parallel conditions/entities. While, as noted, an aim of the QDAMR design is to provide inherent interpretability, the effectiveness with which its outputs serve as explanations for human users remains to be evaluated in future work. 


\small
\bibliographystyle{named}
\bibliography{ijcai22}

\end{document}